\documentclass[11pt]{article}

\usepackage{acl}
\usepackage{times}
\usepackage{latexsym}
\usepackage{enumitem}
\usepackage{xcolor}
\usepackage{booktabs}
\usepackage{graphicx}
\usepackage{multirow}
\usepackage{makecell}
\usepackage{textcomp}
\usepackage[T1]{fontenc}
\usepackage[utf8]{inputenc}
\usepackage{microtype}
\usepackage{inconsolata}
\usepackage{amsmath,amssymb,amsfonts}
\usepackage{tikz}
\usetikzlibrary{arrows.meta,positioning,calc,shapes.geometric}
\usepackage{url}

\newcommand{\HMGCLIP}{\textsc{HMGCLIP}}

\newcommand{\qiuyu}{\color{black}}
\newcommand{\qy}{\color{black}}

\title{HMGCLIP: Heterogeneous Multi-Granularity \\ Contrastive Learning for E-commerce Representation {\qiuyu Learning}}

\author{
\textbf{Qiuyu Zhu} \quad
\textbf{Yi Gao} \quad
\textbf{Zhichao Wan} \quad
\textbf{Mingyang Ma} \\[0.3cm]
{\tt\{zqy527291,xiheng.gy,wanzhichao.wzc,mingyang.mmy\}@alibaba-inc.com}\\
Alibaba International Digital Commerce Group
}

\begin{document}
\maketitle
{
\begin{abstract}

{\qiuyu Although recent Multimodal Large Language Models (MLLMs) have advanced general product understanding, they implicitly encode product information into global embeddings, thereby limiting their ability to capture fine-grained attributes. This limitation hinders performance in tasks requiring precise attribute discrimination, such as distinguishing subtle material differences among visually similar products.} To address this challenge, we propose \HMGCLIP{}, a unified multimodal embedding framework. By constructing a heterogeneous hypergraph, we leverage hypergraph topology to mine structure-aware hard negatives and align multi-granular semantics at both relation and hyperedge levels. This design enables a dual-granularity inference mechanism that dynamically fuses attribute evidence for both fine-grained and coarse-grained downstream tasks. Furthermore, we release a comprehensive fine-grained e-commerce dataset to facilitate future benchmarking. Extensive experiments on this new dataset and the public MAVE benchmark 
show that \HMGCLIP{} outperforms strong multimodal encoders, MLLMs, and e-commerce baselines,
validating the superiority of \HMGCLIP{}. 
\end{abstract}
}


\section{Introduction}
\label{sec:introduction}

Modern e-commerce platforms manage billions of products described by images, titles, structured attributes, and categorical hierarchies. Learning unified representations from heterogeneous signals is fundamental to product search, recommendation, attribute value extraction (PAVE), and duplicate product detection \citep{radford2021learning, chia2022fashionclip}. Despite rapid progress in multimodal representation learning \citep{lin2025mmembed, li2025qwen3vl}, three limitations persist in the e-commerce setting.

\begin{figure}[t]
    \centering
    \includegraphics[width=1.0\columnwidth]{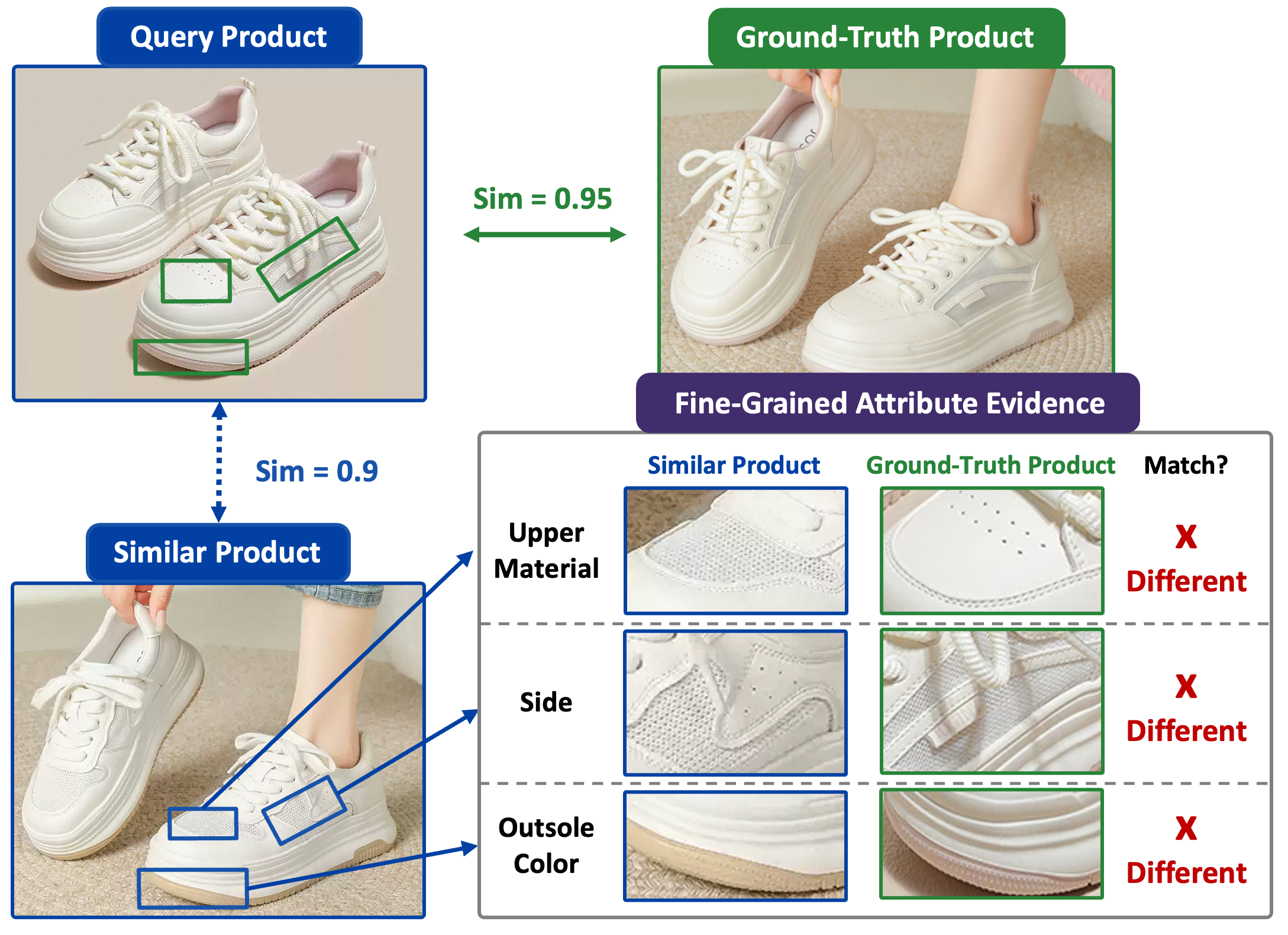}
    \caption{Comparison of fine-grained attributes among the query, ground-truth product and similar product.}
    \label{fig:intro}
\end{figure}

First, general vision-language models such as CLIP \citep{radford2021learning} are trained on web-scale image--text pairs and struggle with the fine-grained distinctions that e-commerce demands---for instance, differentiating between ``mesh'' and ``leather'' uppers, or ``air cushion'' and ``flat rubber'' soles, requires more than global visual alignment. Second, standard contrastive learning typically treats negative samples uniformly or samples them randomly from batches. This approach fails to provide the \emph{semantically confusable hard negatives} necessary for sharpening decision boundaries at the attribute level~\citep{moreira2025nvretriever}. Third, existing benchmarks often lack the structural richness of real-world e-commerce data---particularly category taxonomies, attribute key--value constraints, and higher-order product entity groups---making it difficult to evaluate hierarchical and compositional understanding.

The core challenge is that e-commerce products are inherently \emph{compositional}: their semantics arise from the interaction of visual appearance, textual descriptions, and structured attributes. As illustrated in Figure~\ref{fig:intro}, 
monolithic encoders may fail to preserve fine-grained product distinctions when trained without explicit structural constraints. For example, a query sneaker with a \textit{pink-tinted outsole} and a \textit{mesh upper} may receive high similarity to a visually similar but semantically incorrect item, such as a sneaker with a \textit{beige outsole} and a \textit{leather upper}, because the two products share coarse visual cues including silhouette, lacing pattern, and overall color distribution. In large-scale retrieval, such global bias leads to ranking errors and retrieval ambiguity, as the model overlooks decisive attribute-level evidence such as material texture and outsole appearance.

To address these limitations, we present \textbf{\HMGCLIP{}}, a unified framework for multimodal e-commerce representation learning built on three insights: (1)~e-commerce data naturally forms a \emph{heterogeneous hypergraph} where 
products, images, attributes, and categories participate in higher-order relationships that extend beyond pairwise edges; (2)~effective contrastive learning requires \emph{hard negative mining at multiple granularities}---from coarse category-level to fine attribute-level; {\qiuyu (3)~learning a \emph{structure-aware unified embedding space} enables robust zero-shot generalization across diverse downstream tasks, eliminating the need for task-specific fine-tuning. }
Our main contributions are summarized as:
\begin{itemize}[leftmargin=*]
\item \textbf{Framework.} We present a unified multimodal embedding framework for e-commerce that supports both fine-grained and coarse-grained downstream tasks, achieving robust generalization without task-specific encoder fine-tuning.
\item \textbf{Method.} We develop a multi-granularity contrastive learning paradigm that aligns relation-level and hyperedge-level semantics in a unified embedding space. By exploiting heterogeneous hypergraph topology, our method mines semantically confusable hard negatives and jointly aligns global product representations with local attribute semantics and higher-order groups.
\item \textbf{Dataset.} We release a fine-grained multimodal e-commerce dataset to fill the gap in existing benchmarks and facilitate future research.
\item \textbf{Evaluation.} Extensive experiments on our introduced dataset and the public MAVE benchmark demonstrate that \HMGCLIP{} achieves strong and often state-of-the-art performance across attribute prediction and product classification tasks, validating its robustness and versatility.
\end{itemize}

\section{Related Work}
\label{sec:related_work}

\paragraph{E-commerce Representation Learning.}
Vision-language pre-training models---including CLIP \citep{radford2021learning}, ALIGN \citep{jia2021scalingvisualvisionlanguagerepresentation}, and BLIP-2 \citep{li2023blip2bootstrappinglanguageimagepretraining}---have demonstrated that large-scale image--text alignment yields transferable representations for retrieval and recognition.
This paradigm has been adapted to e-commerce through domain-specific contrastive pre-training and knowledge-enhanced multimodal fusion \citep{dong2022m5productselfharmonizedcontrastivelearning,Zhu_MM2021_KnowledgePerceived,jin2023learninginstancelevelrepresentation}.
Recent MLLM-based e-commerce models improve generative product understanding 
\citep{fu2025moonembedding}, but they are often optimized for task-specific prediction rather than reusable embedding spaces. A parallel line of work addresses product attribute value extraction via generative or instruction-following multimodal LLMs \citep{khandelwal2023largescalegenerativemultimodal, brinkmann2024extractgptexploringpotentiallarge}.
Despite these advances, most methods treat products as isolated image--text pairs and underuse structured dependencies such as attribute-key/value constraints and co-occurring attributes. In contrast, \HMGCLIP{} learns a task-agnostic unified space where products, categories, and attributes are represented as reusable nodes for cross-task generalization.

\paragraph{Contrastive Learning with Hard Negative Mining.}
Prior contrastive learning improves representation quality through label-aware positives, prototypes, graph augmentations, and hard negative mining \citep{khosla2021supervisedcontrastivelearning, li2021prototypicalcontrastivelearningunsupervised,Zhu_WWW2021_GraphContrastive,robinson2021contrastivelearninghardnegative}.
In e-commerce, however, negatives are not arbitrary: they are constrained by category hierarchies, attribute-key/value relations, and visual similarity. Hypergraphs provide a natural abstraction for modeling higher-order product relations beyond pairwise edges. 
\citep{yadati2019hypergcnnewmethodtraining, huang2021unignnunifiedframeworkgraph}, 
and recent e-commerce graph methods have demonstrated the value of structured product relations for retrieval, recommendation, and attribute extraction \citep{wang-etal-2023-fashionklip, hu2025hypergraphbased, hongwimol2026autopkgautomatedframeworkdynamic}. Nevertheless, existing methods rarely unify relation-level hard negatives, hyperedge-level semantic alignment, and multimodal product representations within a single embedding framework. \HMGCLIP{} addresses it by combining heterogeneous hypergraph-based hard negative mining with hierarchical multi-granularity contrastive learning for diverse e-commerce tasks.

\section{Preliminaries}

Products in e-commerce are compositional entities whose semantics arise from heterogeneous fields, including images, text, structured attributes, and category information. For a product $p$, we denote its multimodal inputs as $\mathcal{X}_p = \{I_p, T_p, A_p, C_p\}$, where $I_p$ is the product image, $T_p$ is the title and description, $A_p = \{a_1, a_2, \dots, a_k\}$ is the set of structured aspects, i.e., attribute--value pairs
, and $C_p$ is the category. These modalities capture semantic information at different granularities, ranging from fine-grained physical properties to coarse-grained product roles defined by the platform taxonomy.

Our goal is to learn a unified multimodal embedding function $f(\cdot): \mathcal{X} \rightarrow \mathbb{R}^d$ that captures fine-grained semantics while preserving discriminative boundaries. We evaluate the learned representations on two downstream tasks: attribute prediction~\citep{yang2022mave}, which tests fine-grained discrimination, and product classification, which tests semantic clustering quality.

Following prior works~\citep{yang2022mave, khandelwal2023largescalegenerativemultimodal}, we formulate classification as an embedding-based matching problem. Specifically, given a query product $p$, the predicted attribute value or category is defined as the candidate whose embedding is most semantically similar to the product representation:
\begin{equation}
    y^* = \arg\max_{y \in \hat{\mathcal{Y}}} \mathrm{sim}(f(p), f(y)),
\end{equation}
where $\hat{\mathcal{Y}}$ denotes the candidate set of aspects or categories, and $\text{sim}(\cdot, \cdot)$ represents cosine similarity.

\section{Methodology}

\begin{figure*}[t]
  \centering
  \includegraphics[width=1.0\textwidth]{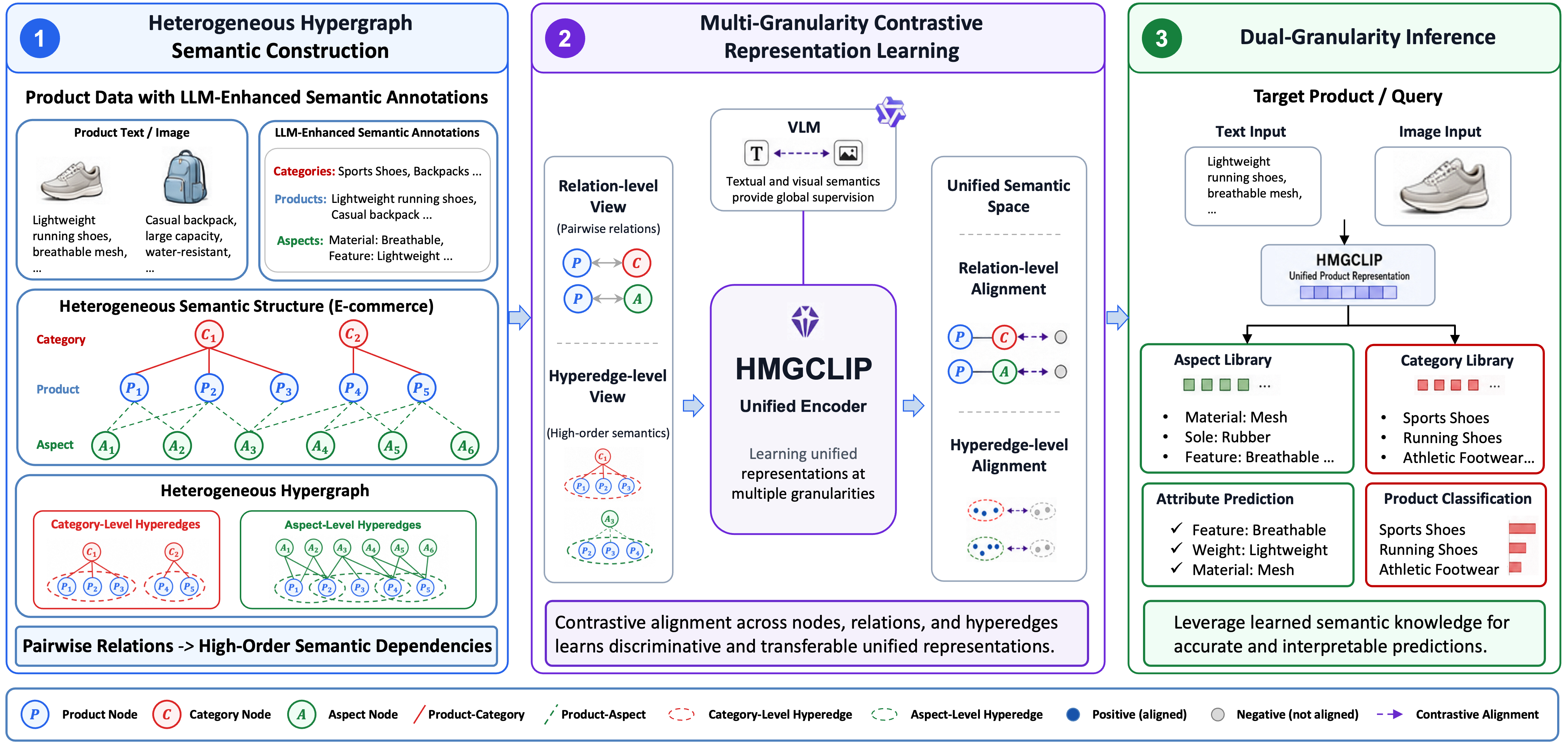}
  \caption{Overview of HMGCLIP. {\qiuyu (1) Construction of a heterogeneous hypergraph from product data and semantic annotations. (2) Multi-granularity contrastive learning over relation-level and hyperedge-level views to unify the semantic space. (3) Dual-granularity inference via retrieval over learned key-value and category libraries for aspect prediction and evidence-fused product classification.}
  }
  \label{fig:framework}
\end{figure*}

As illustrated in Figure~\ref{fig:framework}, HMGCLIP consists of three stages: (1) \textit{Heterogeneous Hypergraph Semantic Construction}, which organizes products, aspects, and categories into a structured semantic graph with pairwise relations and higher-order hyperedges; (2) \textit{Multi-Granularity Contrastive Representation Learning}, which aligns entities at the node, relation, and hyperedge levels to learn a unified semantic space; and (3) \textit{Dual-Granularity Inference}, which performs attribute prediction and evidence-fused product classification by retrieving from the learned aspect and category libraries. HMGCLIP captures both local semantic relations and higher-order semantic dependencies, enabling discriminative representations for downstream e-commerce tasks such as fine-grained attribute prediction and coarse-grained product classification.

\subsection{Heterogeneous Hypergraph Semantic Construction}
\label{subsec:hypergraph_construction}

{\qiuyu We formulate a heterogeneous graph~\cite{wang2019heterogeneous,zhu2025hhgt,zhu2025hierpromptlm} $\mathcal{G} = (\mathcal{V}, \mathcal{E})$ to structure the relationships among products, aspects, and categories, where the node set is defined as $\mathcal{V} = \mathcal{P} \cup \mathcal{A} \cup \mathcal{C}$.} Here, product nodes $\mathcal{P}$ are connected to aspect nodes $\mathcal{A}$ and category nodes $\mathcal{C}$ through product-aspect edges $(p,a) \in \mathcal{E}$ and product-category edges $(p,c) \in \mathcal{E}$, respectively. However, pairwise relations alone cannot capture the group-level semantic consistency in e-commerce catalogs. For example, products sharing the same aspect, such as ``Waterproof,'' often form coherent clusters across different categories. 

{\qiuyu To model such higher-order dependencies, we extend the heterogeneous graph $\mathcal{G}$ into a \textit{heterogeneous hypergraph}~\cite{antelmi2023survey,wang2019heterogeneous}} $\mathcal{G}_H = (\mathcal{V}, \mathcal{E}, \mathcal{H})$, where $\mathcal{H}$ denotes the set of hyperedges. Specifically, we define two types of hyperedges: (1) For each aspect node $a \in \mathcal{A}$, an aspect-level hyperedge $h_a = \{ a \} \cup \{ p \in \mathcal{P} \mid (p, a) \in \mathcal{E} \}$ groups the aspect with all associated products, capturing fine-grained semantic consistency. (2) Likewise, for each category node $c \in \mathcal{C}$, a category-level hyperedge $h_c = \{ c \} \cup \{ p \in \mathcal{P} \mid (p, c) \in \mathcal{E} \}$ groups the category with its member products, preserving coarse-grained semantic coherence. By combining pairwise edges and higher-order hyperedges, $\mathcal{G}_H$ provides a structural foundation for subsequent contrastive learning.

\subsection{Multi-Granularity Contrastive Learning}
\label{subsec:multi_gran_learning}

To learn fine-grained representations, we fine-tune a pre-trained multimodal encoder with e-commerce-specific structural priors through multi-granularity contrastive learning over relation-level and hyperedge-level views. Our method leverages structure-aware sampling to construct informative positive and negative pairs, enabling the model to align pairwise relations and higher-order semantic groups within a unified embedding space.

\subsubsection{Relation-Guided Hard Negative Mining}
\label{subsubsec:hard_neg}

Standard in-batch negatives are often semantically distant and provide limited supervision for fine-grained discrimination. To address this, we mine hard negatives for pairwise relations from the heterogeneous graph. Specifically, for an anchor product $p_i$ and its ground-truth positive aspects $\mathcal{A}_i^+$, we construct a hard negative set $\mathcal{A}_i^-$ by applying three constraints: (i) the candidate aspect must share the same aspect key as a positive aspect, such as ``Material'' or ``Color''; (ii) it must appear in the same category context as $p_i$; and (iii) it must not be a ground-truth aspect of $p_i$. This yields semantically confusable negatives for the product-aspect relation,
encouraging the model to learn sharper distinctions among directly related aspects.

\subsubsection{Hyperedge-Guided Group Alignment}
\label{subsubsec:hyperedge}


{\qiuyu Beyond pairwise supervision, we enforce group-level consistency via hyperedge alignment. For each anchor product, we define positives as co-occurring products within the same hyperedge and negatives as those outside it. This contrastive objective maximizes anchor-positive similarity while minimizing anchor-negative similarity, fostering compact and coherent clustering of semantically related products in the latent space.}

\subsection{Joint Relation-Level and Hyperedge-Level Contrastive Optimization}

To jointly capture fine-grained pairwise semantics and higher-order group structure, we optimize a unified contrastive objective over relation-level and hyperedge-level supervision. Specifically, we adopt the InfoNCE loss~\cite{oord2018representation} as the basic optimization framework. 

For an anchor sample $i$, let $\mathcal{S}_i^+$ and $\mathcal{S}_i^-$ denote the sets of positive and negative samples, respectively. We define the positive and negative partition functions as:
\begin{equation}
    Z_i^+ = \sum_{j \in \mathcal{S}_i^+} \exp(\mathbf{h}_i^\top \mathbf{h}_j / \tau), 
\end{equation}
\begin{equation}
    Z_i^- = \sum_{k \in \mathcal{S}_i^-} \exp(\mathbf{h}_i^\top \mathbf{h}_k / \tau),
\end{equation}
where $\tau$ is the temperature parameter. The loss for a batch of $N$ samples is formulated as:
\begin{equation}
    \mathcal{L}_{\text{contrast}} = -\frac{1}{N} \sum_{i=1}^{N} \log \frac{Z_i^+}{Z_i^+ + Z_i^-}.
\end{equation}

We apply this loss to both relation-level and hyperedge-level views. Specifically, $\mathcal{L}_{rel}$ is computed using product-aspect pairs with hard negatives, while $\mathcal{L}_{hyper}$ is computed using product groups within semantic hyperedges defined in Section~\ref{subsec:hypergraph_construction}. The overall objective is defined as:
\begin{equation}
    \mathcal{L} = \mathcal{L}_{rel} + \lambda \mathcal{L}_{hyper},
\end{equation}
where $\lambda$ balances the contribution of the two supervision signals.

\subsection{Dual-Granularity Inference}
\label{subsec:inference}

Upon completion of multi-granularity contrastive pre-training, the encoders are frozen. We propose a dual-path inference mechanism that adapts to task granularity by leveraging the aligned embedding space for both fine-grained aspect retrieval and coarse-grained category prediction.

\subsubsection{Path I: {\qiuyu Fine-Grained Retrieval}}

For fine-grained aspect prediction, we directly utilize the multimodal embedding $\mathbf{z}_q$ of the query product $q$ to retrieve the best-matching aspect from the vocabulary $\mathcal{A}$. The predicted aspect $\hat{a}_q$ is identified by maximizing the cosine similarity between the query and candidate aspect embeddings:
\begin{equation}
    \hat{a}_q = \arg\max_{j \in \mathcal{A}} \frac{\mathbf{z}_q^\top \mathbf{a}_j}{\|\mathbf{z}_q\| \|\mathbf{a}_j\|},
\end{equation}
where $\mathbf{a}_j$ denotes the embedding of aspect candidate $j$. This non-parametric nearest-neighbor search leverages the discriminative embedding space to capture fine-grained semantic distinctions without requiring additional classification heads.

\subsubsection{Path II: Evidence-Fused Coarse-Grained Retrieval}
\label{subsec:semantic_fusion}



For coarse-grained category retrieval, relying solely on the product's intrinsic features may lead to ambiguity in distinguishing semantically similar categories. To address this, we construct an \textit{enhanced query representation} by integrating specific semantic evidence into the global context.

\paragraph{Semantic Anchor-Based Fusion.}
Initially, we utilize the set of top-$K$ aspects retrieved in Path I to construct a robust semantic anchor. Let $\{\mathbf{a}_1, \mathbf{a}_2, \ldots, \mathbf{a}_K\}$ be the embeddings of these retrieved aspects. We obtain the aggregated aspect representation $\mathbf{a}_{agg}$ via mean pooling. This aggregated vector serves as a contextual summary of the product's key aspects. We then fuse the product's intrinsic multimodal embedding $\mathbf{z}_q$ with this semantic summary via linear interpolation:
\begin{equation}
    \mathbf{z}_{fused} = \alpha \mathbf{z}_q + (1-\alpha) \mathbf{a}_{agg},
\end{equation}
where $\alpha$ balances the contribution of intrinsic features and aggregated semantic evidence. The final category prediction $\hat{y}_q$ is obtained by retrieving the best match from the category candidate set $\mathcal{C}$ based on cosine similarity:
\begin{equation}
    \hat{y}_q = \arg\max_{c \in \mathcal{C}} \, \frac{(\mathbf{z}_{fused})^\top \mathbf{c}}{\|\mathbf{z}_{fused}\| \|\mathbf{c}\|}.
\end{equation}


\paragraph{Residual Transformer-based Fusion.}
{To capture fine-grained semantic interactions between product features and aspect evidence, we further introduce an evidence-enhanced residual fusion module. This module injects retrieved aspect evidence into the product representation via a residual connection, thereby enriching semantic details while preserving the geometric structure of the learned embedding space. 


Given the product embedding $\mathbf{z}_q$ and the set of retrieved aspect embeddings $\{\mathbf{a}_1, \dots, \mathbf{a}_K\}$ defined previously, we form a field-level token sequence by combining the product token with its aspect evidence tokens: $\mathbf{X} = [\mathbf{z}_q, \mathbf{a}_1, \mathbf{a}_2, \ldots, \mathbf{a}_K]$. Then, we add learnable type embeddings to distinguish the product token from aspect tokens:
\begin{equation}
    \tilde{\mathbf{x}}_0 = \mathbf{z}_q + \mathbf{e}_{prod},
\end{equation}
\begin{equation}
    \tilde{\mathbf{x}}_i = \mathbf{a}_i + \mathbf{e}_{attr}, \quad i=1,\ldots,K,
\end{equation}
where $\mathbf{e}_{prod}$ and $\mathbf{e}_{attr}$ are learnable type embeddings. The resulting sequence is denoted as: $\tilde{\mathbf{X}} = [\tilde{\mathbf{x}}_0,\tilde{\mathbf{x}}_1,\ldots,\tilde{\mathbf{x}}_K]$.
After that, we apply a lightweight Transformer encoder to model interactions between the product representation and the retrieved aspect evidence:
\begin{equation}
    [\mathbf{h}_0,\mathbf{h}_1,\ldots,\mathbf{h}_K]
    = \mathrm{TransformerEncoder}(\tilde{\mathbf{X}}).
\end{equation}
Through self-attention, the product token can selectively attend to category-relevant aspects, while aspect tokens can also contextualize one another. This enables interaction-aware fusion beyond mean pooling or scalar weighting.


Instead of directly replacing the product embedding with the transformed token 
$\mathbf{h}_0$, we use it to predict a residual correction, and obtain the final
representation through residual composition:
\begin{equation}
\mathbf{z}
= \operatorname{norm}\!\left({\qy \mathbf{z}_q} + \alpha_r f_{\theta}(\mathbf{h}_0)\right).
\label{eq:residual_composition}
\end{equation}
where $f_{\theta}(\cdot)$ is a lightweight projection module, $\alpha_r$ controls the aspect-aware correction strength, and $\mathrm{norm}(\cdot)$ denotes L2 normalization. This residual design keeps the original product embedding as the semantic anchor while allowing retrieved aspect evidence to provide a controlled correction. Thus, the fused representation remains compatible with the learned embedding space and can be directly matched against taxonomy-defined categories.

\begin{table*}[t]
\centering
\scriptsize
\setlength{\tabcolsep}{3pt}
\renewcommand{\arraystretch}{0.88}
\caption{{\qy Performance of aspect (attribute) prediction and product classification tasks on both datasets.
}}
\label{tab:main_results_internal}
\resizebox{\textwidth}{!}{
\begin{tabular}{lccccc|ccccc}
\toprule
\multirow{2}{*}{\textbf{Model}}
& \multicolumn{5}{c|}{\textbf{Internal Dataset}}
& \multicolumn{5}{c}{\textbf{MAVE}} \\
\cmidrule(lr){2-6} \cmidrule(lr){7-11}
& \textbf{Hit@1}/\textbf{MRR@1}
& \textbf{Hit@3} & \textbf{MRR@3}
& \textbf{Hit@5} & \textbf{MRR@5}
& \textbf{Hit@1}/\textbf{MRR@1}
& \textbf{Hit@3} & \textbf{MRR@3}
& \textbf{Hit@5} & \textbf{MRR@5} \\
\midrule
\addlinespace[0.5em]
\multicolumn{11}{c}{\textbf{Aspect Prediction}} \\
\midrule
SigLIP2
& 40.74 & 76.27 & 56.64 & 87.68 & 59.26
& 5.73 & 15.60 & 9.90 & 24.42 & 11.90 \\

FashionCLIP
& 49.63 & 81.59 & 64.00 & 90.54 & 66.06
& \underline{16.71} & \underline{36.00} & \underline{25.00} & \underline{46.92} & \underline{27.48} \\
\midrule
InternVL3.5-2B
& 35.29 & 72.09 & 51.59 & 83.53 & 54.25
& 5.29 & 14.90 & 9.31 & 24.30 & 11.43 \\

Qwen3-VL-2B
& 37.88 & 72.86 & 53.46 & 84.00 & 56.03
& 6.01 & 14.68 & 9.64 & 21.68 & 11.20 \\
\midrule
GME-Qwen2VL
& 40.61 & 75.58 & 56.13 & 86.48 & 58.65
& 11.40 & 26.76 & 17.98 & 37.32 & 20.38 \\

MM-Embed
& 47.09 & 79.06 & 61.32 & 89.65 & 63.75
& 5.48 & 15.93 & 9.87 & 23.43 & 11.57 \\

CASLIE-S
& 39.19 & 74.73 & 54.99 & 84.94 & 57.35
& 8.13 & 20.93 & 13.60 & 29.94 & 15.66 \\

Qwen3-VL-Emb
& \underline{54.10} & \underline{85.27} & \underline{68.12} & \underline{92.74} & \underline{69.86}
& 15.56 & 33.33 & 23.25 & 45.38 & 25.97 \\
\midrule
\HMGCLIP{}
& \textbf{75.38} & \textbf{94.70} & \textbf{84.27} & \textbf{97.61} & \textbf{84.95}
& \textbf{24.61} & \textbf{46.16} & \textbf{33.93} & \textbf{58.32} & \textbf{36.69} \\
\midrule
\addlinespace[0.5em]
\multicolumn{11}{c}{\textbf{Product Classification}} \\
\midrule
SigLIP2
& 0.17 & 0.41 & 0.28 & 0.72 & 0.35
& 1.05 & 1.18 & 1.11 & 1.23 & 1.12 \\

FashionCLIP
& 39.79 & 60.53 & 48.94 & 68.51 & 50.76
& \underline{75.66} & \underline{85.33} & \underline{80.03} & 89.20 & \underline{80.89} \\
\midrule
InternVL3.5-2B
& 6.81 & 15.36 & 10.47 & 21.73 & 11.92
& 8.99 & 40.24 & 24.04 & 48.27 & 25.86 \\

Qwen3-VL-2B
& 0.27 & 0.78 & 0.48 & 1.24 & 0.59
& 22.49 & 34.89 & 27.99 & 37.87 & 28.68 \\
\midrule
GME-Qwen2VL
& 24.25 & 45.14 & 33.36 & 55.68 & 35.78
& 14.73 & 25.18 & 19.12 & 32.51 & 20.79 \\

MM-Embed
& 2.58 & 4.69 & 3.48 & 6.43 & 3.87
& 9.54 & 12.54 & 10.83 & 14.37 & 11.25 \\

CASLIE-S
& 9.46 & 19.45 & 13.72 & 25.38 & 15.07
& 12.32 & 42.69 & 26.10 & 50.97 & 28.01 \\

Qwen3-VL-Emb
& \underline{48.30} & \underline{71.19} & \underline{58.46} & \underline{79.79} & \underline{60.45}
& 61.45 & 83.63 & 71.44 & \underline{89.83} & 72.87 \\
\midrule
\HMGCLIP{}
& \textbf{84.23} & \textbf{93.77} & \textbf{88.55} & \textbf{95.75} & \textbf{89.01}
& \textbf{96.70} & \textbf{99.15} & \textbf{97.83} & \textbf{99.57} & \textbf{97.93} \\
\bottomrule
\end{tabular}
}
\end{table*}

\section{Experiments and Analysis}
\label{sec:experiments}

\subsection{Datasets}
\label{subsec:datasets}

\noindent\textbf{MAVE~\citep{yang2022mave}.} From the original 2.2M products, we retain multimodal products with images and structured aspects, prioritizing items with at least three aspects and adding two-aspect samples under a per-category cap to reduce imbalance. 

\noindent\textbf{Internal dataset.} We construct a dataset derived from real-world product listings on a Southeast Asian e-commerce platform. The data encompasses titles, descriptions, {\qy images}, category labels, and structured key–value pairs. Reflecting the region's linguistic diversity, the dataset is multilingual, predominantly featuring English alongside Thai, Vietnamese, Indonesian, and Chinese. It is specifically designed to address two core challenges: coarse-grained category understanding and fine-grained aspect discrimination. 



\subsection{Baselines and Evaluation Metrics}
\label{subsec:baselines}

All methods are evaluated under the same candidate pools. We report Hit@$K$ and MRR@$K$ for both fine-grained aspect prediction and coarse-grained product classification. For our method, Qwen3-VL-Embedding-2B serves as the backbone and is further post-trained with heterogeneous multi-granularity contrastive learning. At inference, aspect prediction uses direct product-to-aspect matching, while category prediction adopts aspect-guided semantic fusion (\S\ref{subsec:semantic_fusion}). 

\subsection{Experimental Results}

\begin{table*}[thb]
\centering
\scriptsize
\setlength{\tabcolsep}{10pt}
\renewcommand{\arraystretch}{1.0}
\caption{Qualitative comparison on aspect (attribute) prediction. \textcolor{green!50!black}{$\checkmark$} denotes correct Top-1; \textcolor{red}{$\times$} denotes incorrect.}
\label{tab:qualitative_pave}
\resizebox{\textwidth}{!}{
\begin{tabular}{cm{1.3cm}p{3.0cm}p{2.2cm}p{2.2cm}p{2.2cm}}
\toprule
\# & Image & Product & Ground Truth & Qwen3-VL-Emb & \HMGCLIP{} \\
\midrule
1 & \raisebox{-0.5\height}{\includegraphics[height=1.1cm]{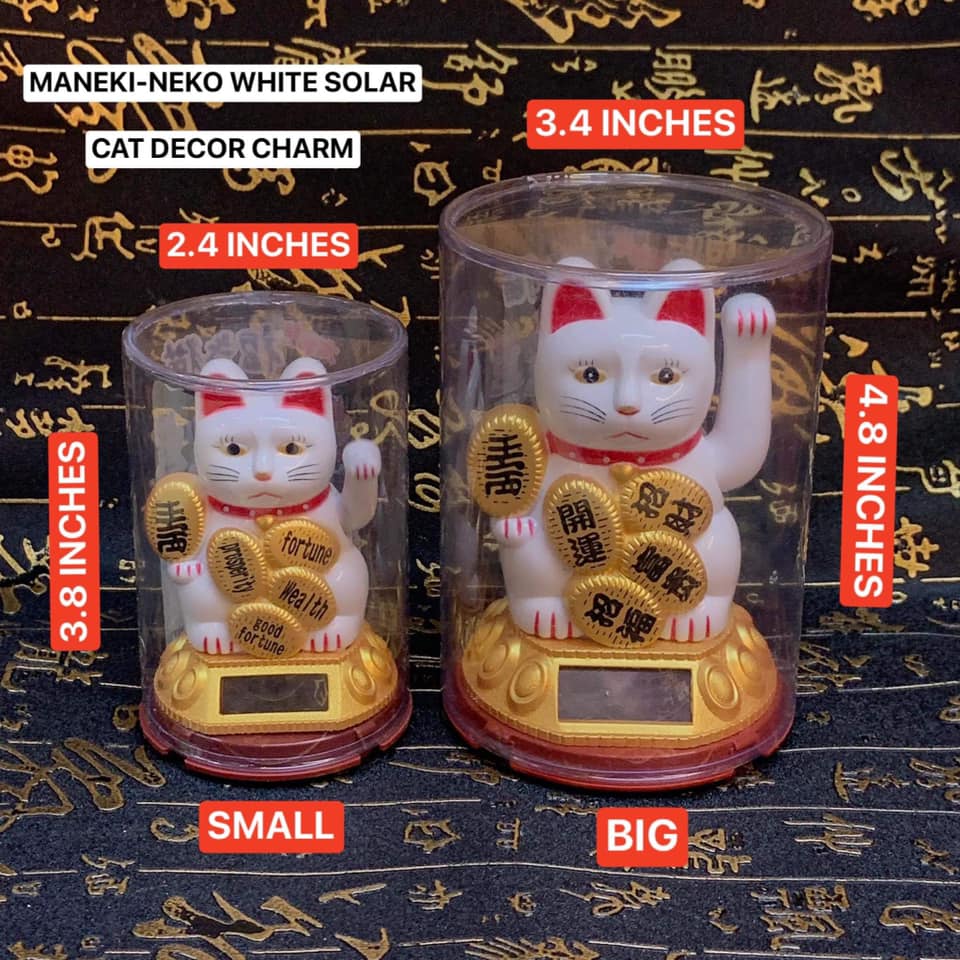}} & PIN PIN LUCKY CHARM MANEKI-NEKO WHITE SOLAR CAT DECOR CHARM & material: plastic & \textcolor{red}{$\times$} material: brass & \textcolor{green!50!black}{$\checkmark$} material: plastic \\
2 & \raisebox{-0.5\height}{\includegraphics[height=1.1cm]{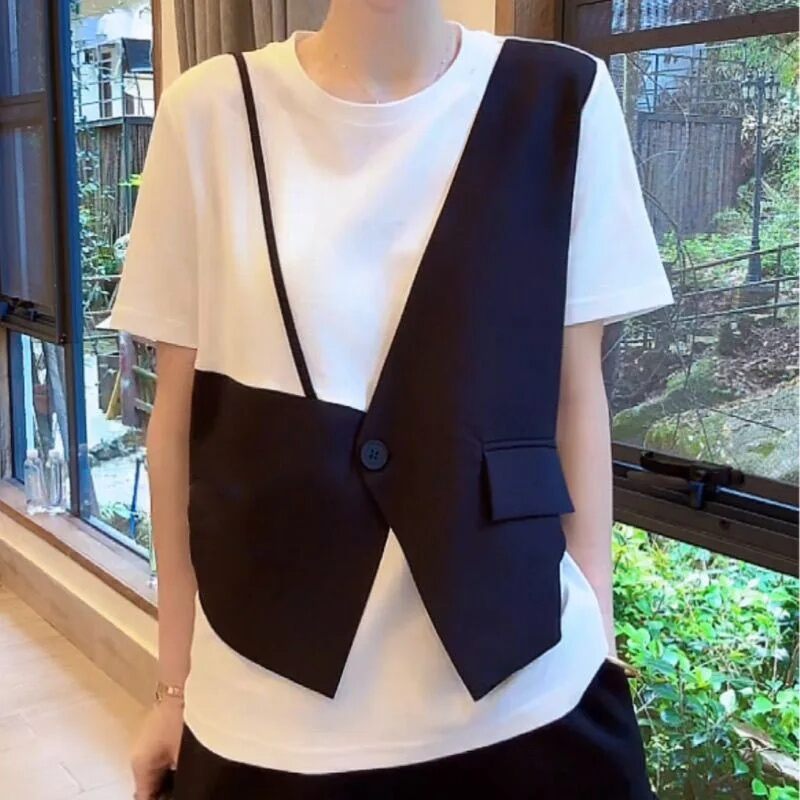}} & Summer New Fashion Loose Womens T-shirt Short Sleeve Patchwork Vest Top Bottomin\dots & clothing decoration: button & \textcolor{red}{$\times$} clothing decoration: side slit & \textcolor{green!50!black}{$\checkmark$} clothing decoration: button \\
3 & \raisebox{-0.5\height}{\includegraphics[height=1.1cm]{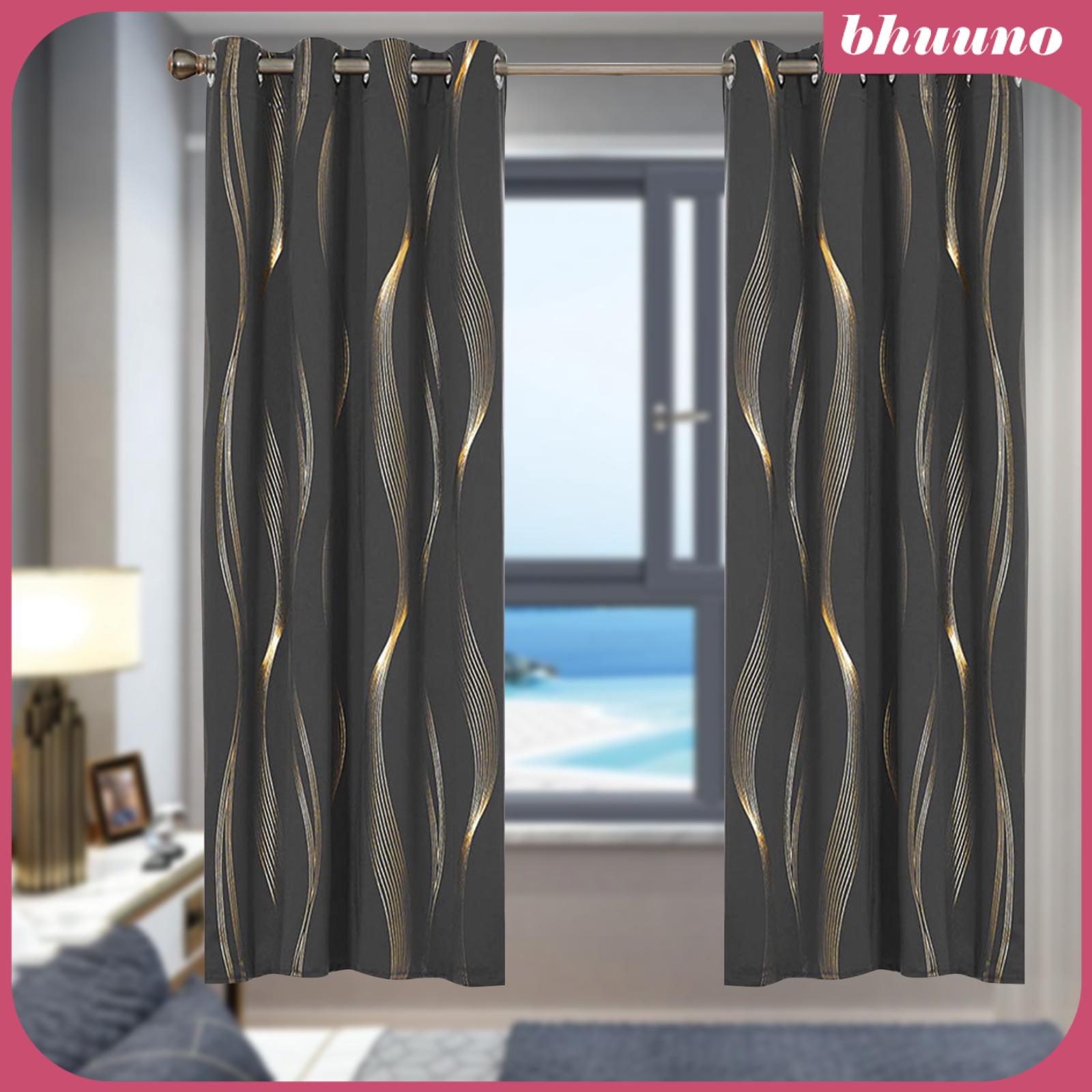}} & Bhuuno Grommet r\`em cua so R\`em cua so Phong C\'{a}ch Trang Tr\'{i} N\d{o}i Th\'{a}t Tra\dots & curtain material: polyester & \textcolor{red}{$\times$} curtain material: vinyl & \textcolor{green!50!black}{$\checkmark$} curtain material: polyester \\
\bottomrule
\end{tabular}
}
\end{table*}

\begin{table*}[thb]
\centering
\scriptsize
\setlength{\tabcolsep}{10pt}
\renewcommand{\arraystretch}{1.0}
\caption{Qualitative comparison on category classification. \textcolor{green!50!black}{$\checkmark$} denotes correct Top-1; \textcolor{red}{$\times$} denotes incorrect.}
\label{tab:qualitative_cate}
\resizebox{\textwidth}{!}{
\begin{tabular}{cm{1.3cm}p{3.0cm}p{2.2cm}p{2.2cm}p{2.2cm}}
\toprule
\# & Image & Product & Ground Truth & Qwen3-VL-Emb & \HMGCLIP{} \\
\midrule
1 & \raisebox{-0.5\height}{\includegraphics[height=1.1cm]{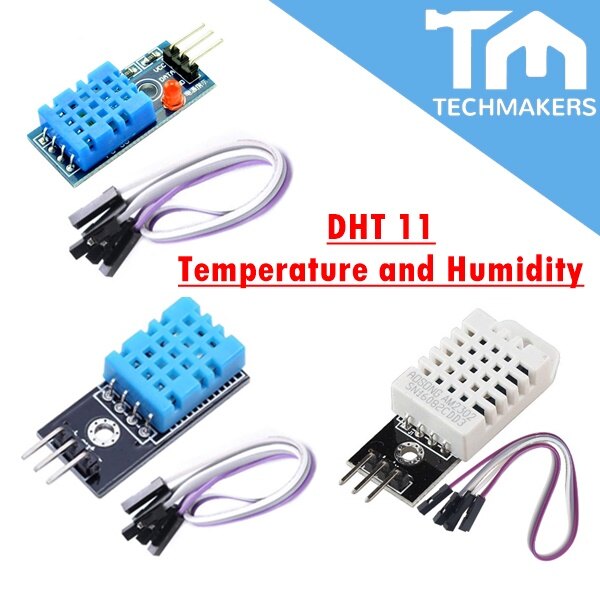}} & DHT11 DHT 11 High Accuracy Temperature and Humidity Moisture Sensor Detect 3.3V 5V Module\dots & Tools \& Home Improvement > Electrical > Electrical Circuitry \& Parts & \textcolor{red}{$\times$} Home Appliances > Heating, Cooling and Ventilation > Humidifiers & \textcolor{green!50!black}{$\checkmark$} Tools \& Home Improvement > Electrical > Electrical Circuitry \& Parts \\
2 & \raisebox{-0.5\height}{\includegraphics[height=1.1cm]{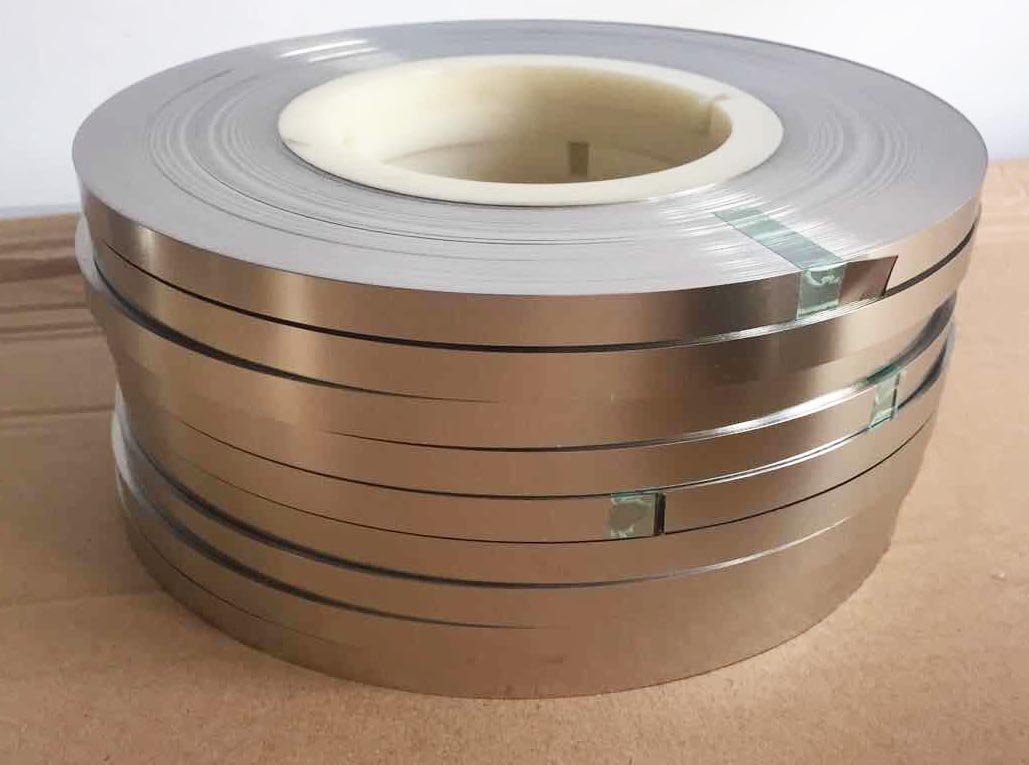}} & K\d{e}m h\`an cell pin cu\d{o}n \dj on 1 m\'{e}t k\d{e}m k\'{i}ch thu\'{o}c 0.15*8mm & Tools \& Home Improvement > Power Tools \& Accessories > Power Tools Parts & \textcolor{red}{$\times$} Stationery, Craft \& Gift Cards > Packaging \& Cartons > Packaging Protection > Shrink Wrap & \textcolor{green!50!black}{$\checkmark$} Tools \& Home Improvement > Power Tools \& Accessories > Power Tools Parts \\
3 & \raisebox{-0.5\height}{\includegraphics[height=1.1cm]{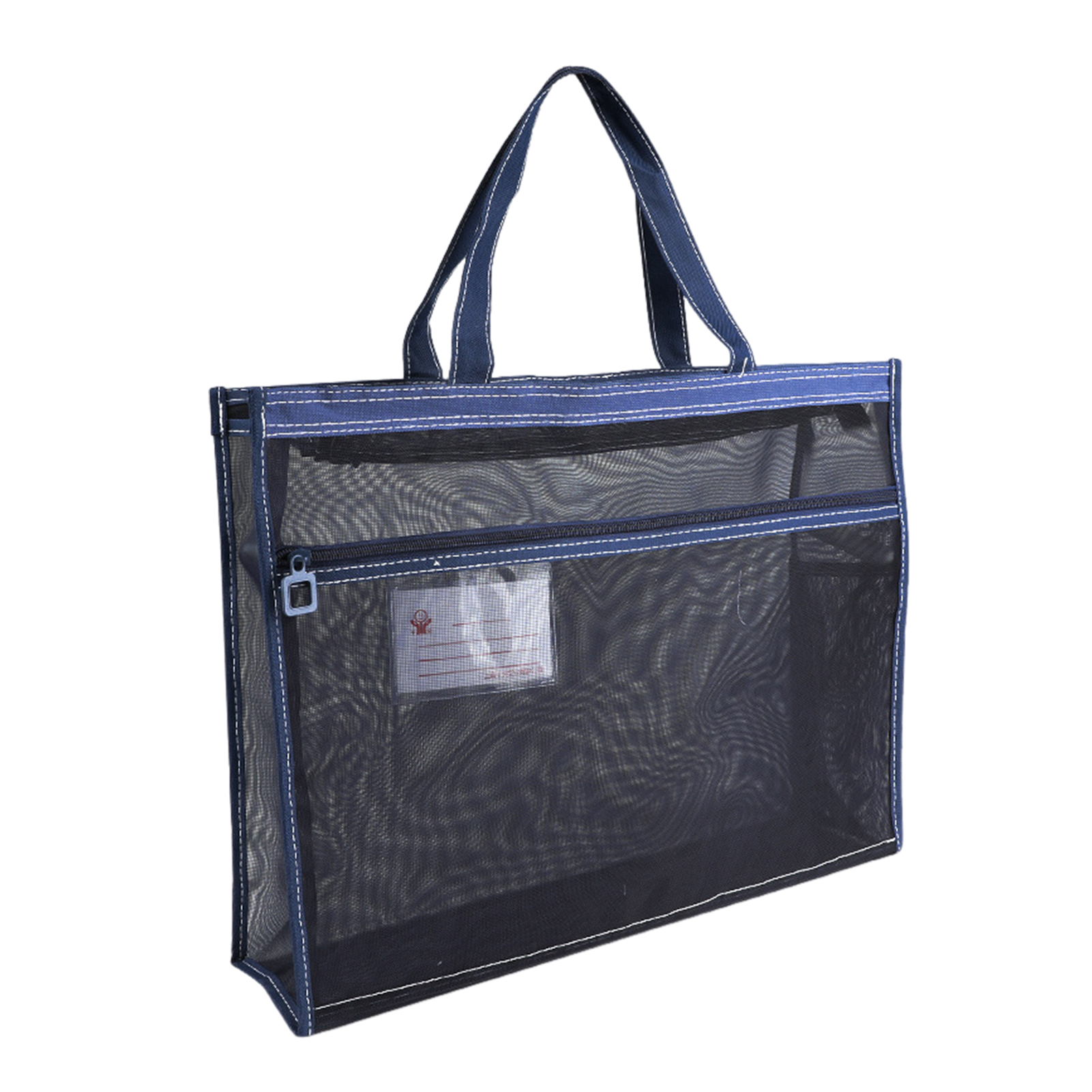}} & Mesh Art Supplies Bag Portable Art Supplies Organizer Large A3 Size Me\dots & Stationery, Craft \& Gift Cards > School \& Office > Storage \& Organizers > Filing \& Document Presentation & \textcolor{red}{$\times$} Bags and Travel > Functional Bags > Library Bag & \textcolor{green!50!black}{$\checkmark$} Stationery, Craft \& Gift Cards > School \& Office > Storage \& Organizers > Filing \& Document Presentation \\
\bottomrule
\end{tabular}
}
\end{table*}

{\qy To evaluate the performance of \HMGCLIP{}, we compare it with all baselines on both tasks across both datasets. The results are presented in Table~\ref{tab:main_results_internal}.

\noindent\textbf{Aspect Prediction.}
\HMGCLIP{} outperforms all baselines, demonstrating superior fine-grained semantic capture. On the internal dataset, it achieves a Hit@1/MRR@1 of 75.38\%, surpassing the strongest baseline (Qwen3-VL-Emb) by 21.28\%. This margin highlights its enhanced ability to distinguish subtle aspect differences compared to general-purpose VLMs. The advantage persists across broader recall scopes, with a Hit@5 of 97.61\% (nearly 5\% higher than the second-best model). Furthermore, \HMGCLIP{} generalizes well to the heterogeneous MAVE benchmark, achieving a Hit@1/MRR@1 of 24.61\% and MRR@5 of 36.69\%. By exceeding specialized models like FashionCLIP (Hit@1: 16.71\%), our approach proves robust in capturing complex product-aspect correlations in open-world scenarios. In particular, graph-derived hard negatives expose the model to semantically confusable attribute values, leading to sharper product--aspect decision boundaries.

\noindent\textbf{Product Classification.}
\HMGCLIP{} consistently dominates baselines in coarse-grained category distinction. On the internal dataset, it achieves a Hit@1 of 84.23\%, exceeding the second-best model (Qwen3-VL-Emb, 48.30\%) by over 35\%. On the MAVE benchmark, it attains a Hit@1 of 96.70\% and Hit@5 of 99.57\%, outperforming FashionCLIP (Hit@1: 75.66\%). These results validate our \textit{Dual-Granularity Inference} built on \textit{Residual Transformer Fusion}. The low performance of SigLIP2 is likely due to its pairwise sigmoid image--text objective and lack of explicit cross-modal fusion. By preserving critical semantic signals during multimodal integration, this architecture creates a robust feature space that reinforces category boundaries through aspect-level consistency.

\subsection{{\qy Analysis of Evidence Fusion Strategies}}

We study evidence fusion through a progressive design axis, ranging from uniform aggregation to anchor-preserving weighting, instance-adaptive weighting, and interaction-aware residual fusion for coarse-grained category retrieval.

First, {\qy aspect} evidence must be incorporated carefully: mean pooling slightly improves Hit@1 over the backbone, but degrades Hit@3 and Hit@5, suggesting that uniform aggregation can disturb the neighborhood structure of the product embedding space. 
Second, preserving the product representation as a semantic anchor is important. 
Manually increasing the product-token weight improves over mean pooling, indicating that aspects should correct rather than replace the global product representation. 
Third, instance-adaptive weighting further improves performance, showing that different products rely on different subsets of evidence.

{\qy
\subsection{Qualitative Analysis}
\label{subsec:qualitative}

To complement our quantitative results, we present qualitative comparisons in Table~\ref{tab:qualitative_pave} and Table~\ref{tab:qualitative_cate}, illustrating how \HMGCLIP{} rectifies errors made by the strongest baseline, Qwen3-VL-Emb.

\noindent\textbf{Aspect Prediction.}
Table~\ref{tab:qualitative_pave} presents the Hit@1 results, illustrating that \HMGCLIP{} correctly retrieves the ground truth for semantically ambiguous fine-grained aspects where the baseline fails. For instance, it distinguishes material differences (``plastic'' vs. ``brass'' for a charm; ``polyester'' vs. ``vinyl'' for curtains) and structural details (``button'' vs. ``side slit'' on a T-shirt). These results validate the efficacy of our heterogeneous multi-granularity contrastive learning, which enhances fine-grained discriminability by aligning visual features with specific aspect granularities, thereby sharpening decision boundaries among confusable values.

\noindent\textbf{Product Classification.}
Table~\ref{tab:qualitative_cate} demonstrates the advantage of our evidence-fused coarse-grained retrieval. While Qwen3-VL-Emb misclassifies items into semantically overlapping but functionally distinct categories (e.g., DHT11 sensor as ``Humidifiers,'' welding tape as ``Shrink Wrap''), \HMGCLIP{} accurately predicts specific sub-categories (``Electrical Circuitry \& Parts,'' ``Power Tools Parts''). This improvement stems from our residual transformer fusion mechanism, which injects retrieved aspect evidence into the classification process. By leveraging this fused evidence, the model effectively disambiguates products sharing general contextual features, reinforcing category boundaries through aspect-level consistency.
}

\section{Conclusion}
\label{sec:conclusion}


In this paper, we present \HMGCLIP{}, a unified multimodal embedding framework for multi-granularity e-commerce representation learning. By leveraging a heterogeneous hypergraph as a structural prior, \HMGCLIP{} establishes a multi-granularity contrastive learning paradigm that aligns relation-level and hyperedge-level semantics within a unified embedding space. This design enables a dual-granularity inference mechanism that achieves generalization across both fine-grained and coarse-grained tasks without requiring task-specific fine-tuning. We further release a comprehensive multimodal e-commerce dataset. Experiments on both datasets demonstrate that \HMGCLIP{} achieves state-of-the-art performance in both tasks, validating its effectiveness and versatility.

\bibliography{main}

\end{document}